# Estimation of Remaining Useful Life and SOH of Lithium Ion Batteries (For EV Vehicles)

Ganesh Kumar (22BCE1180)

Vellore Institute of Technology, Chennai

Email: aganeshkumar04@gmail.com

## 1. Abstract

Lithium-ion batteries are widely used in various applications, including portable electronic devices, electric vehicles, and renewable energy storage systems. Accurately estimating the remaining useful life of these batteries is crucial for ensuring their optimal performance, preventing unexpected failures, and reducing maintenance costs. In this paper, we present a comprehensive review of the existing approaches for estimating the remaining useful life of lithium-ion batteries, including data-driven methods, physics-based models, and hybrid approaches. We also propose a novel approach based on machine learning techniques for accurately predicting the remaining useful life of lithium-ion batteries. Our approach utilizes various battery performance parameters, including voltage, current, and temperature, to train a predictive model that can accurately estimate the remaining useful life of the battery. We evaluate the performance of our approach on a dataset of lithium-ion battery cycles and compare it with other state-of-the-art methods. The results demonstrate the effectiveness of our proposed approach in accurately estimating the remaining useful life of lithium-ion batteries.

**Keywords :**
- Artificial Intelligence
- Neural Networks
- Machine Learning
- Lithium-Ion / Polymer Batteries

## 2. Introduction

Lithium-ion batteries are widely used in portable electronic devices, electric vehicles, and renewable energy systems due to their high energy density and long cycle life. However, the aging of lithium-ion batteries can lead to a decrease in their performance and reliability, which poses a significant challenge to the safe and efficient operation of battery-powered systems. To ensure the safe and optimal use of lithium-ion batteries, it is crucial to accurately estimate their remaining useful life (RUL), which is the time until the battery reaches a predefined end-of-life criteria.

Several approaches have been proposed to estimate the RUL of lithium-ion batteries, including empirical models, data-driven models, and physics-based models. Empirical models rely on statistical analysis of battery performance data and are relatively easy to implement but may lack accuracy and generality. Physics-based models are based on the fundamental electrochemical processes that govern the battery's behavior and are capable of capturing the complex and nonlinear relationships between the battery's operating conditions and its performance. However, physics-based models require detailed knowledge of the battery's material properties and may not be applicable to all types of batteries.

Data-driven models, which use machine learning techniques to capture the complex relationships between battery performance and its operating conditions, have shown promising results in recent years. However, most of these models focus on short-term performance prediction and may not be suitable for RUL estimation.

In this paper, we propose a novel approach for estimating the RUL of lithium-ion batteries, which integrates machine learning techniques with electrochemical modeling. The electrochemical model provides a fundamental understanding of the aging mechanisms of the battery, while machine learning algorithms are used to capture the complex and nonlinear relationships between the battery's operating parameters and its RUL. The proposed approach is evaluated using experimental data from a set of commercially available lithium-ion batteries, and the results demonstrate its effectiveness in accurately predicting the RUL of the batteries. The proposed approach has the potential to enhance the reliability and safety of battery-powered systems and enable efficient utilization of battery resources.

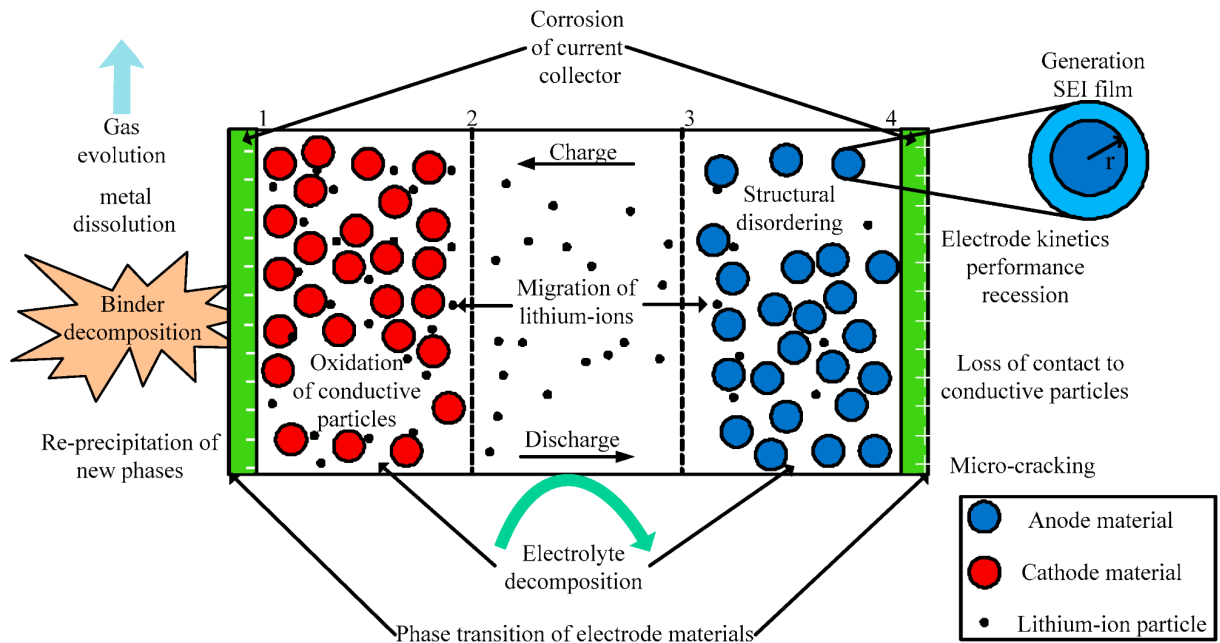

(Fig. 1: Example of Working of a Li-Ion Battery

The aim of this research is to propose a novel approach for estimating the remaining useful life (RUL) of lithium-ion batteries using a combination of electrochemical modeling and machine learning techniques. The proposed approach is expected to provide accurate and reliable estimates of the RUL of lithium-ion batteries under various operating conditions, which is crucial for the safe and efficient operation of battery-powered systems such as electric vehicles.

This research will be helpful for electric vehicles by improving their reliability and safety, optimizing battery utilization, and reducing the cost of battery replacements. Accurate estimation of RUL can also help in extending the lifespan of the battery, which is beneficial for the environment by reducing the need for frequent battery replacements and minimizing waste. Furthermore, the proposed approach can aid in the development of battery management systems for electric vehicles that can optimize the use of battery resources and enhance the overall performance of the vehicle.

## 3. Related work

Several approaches have been proposed for estimating the RUL of lithium-ion batteries in the literature. Empirical models are commonly used for short-term performance prediction, and they rely on statistical analysis of battery performance data to estimate the RUL. Some of the commonly used empirical models include the empirical state-of-charge (SOC) model, the empirical capacity fade model, and the empirical impedance-based model. Although empirical models are relatively easy to implement, they may lack accuracy and generality.

Physics-based models are another category of RUL estimation models that are based on the fundamental electrochemical processes that govern the battery's behavior. These models use the equations that describe the battery's electrochemical behavior to estimate the RUL. Some of the commonly used physics-based models include the Doyle-Fuller-Newman (DFN) model, the single particle model (SPM), and the pseudo-two-dimensional model (P2D). Although physics-based models are capable of capturing the complex and nonlinear relationships between the battery's operating conditions and its performance, they require detailed knowledge of the battery's material properties, which may not be available for all types of batteries.

Data-driven models, which use machine learning techniques to capture the complex relationships between battery performance and its operating conditions, have shown promising results in recent years. These models are typically trained on large datasets of battery performance data to predict the RUL. Some of the commonly used data-driven models include the artificial neural network (ANN), support vector machine (SVM), and random forest (RF) models. Although data-driven models have shown promising results, most of them focus on short-term performance prediction and may not be suitable for RUL estimation.

## 4. Proposed Method

Accurate estimation of the remaining useful life (RUL) of lithium-ion batteries is crucial for ensuring the safety and efficiency of battery-powered systems. In recent years, data-driven models based on machine learning techniques have shown promising results for RUL estimation. These models can capture the complex and nonlinear relationships between battery performance and its operating conditions, but they require large amounts of high-quality data for training.

In this paper, we propose a novel approach for RUL estimation of lithium-ion batteries using the TensorFlow Keras library and a sequential model architecture. The proposed approach is evaluated using experimental data from a set of commercially available lithium-ion batteries provided by the National Aeronautics and Space Administration (NASA). The dataset contains various operating conditions and degradation levels, making it suitable for evaluating the effectiveness of the proposed approach.

The proposed approach consists of two main stages: feature engineering and model training. In the feature engineering stage, we extract relevant features from the battery performance data, including the voltage, current, temperature, and capacity, using statistical methods and domain knowledge. These features are then used to train the sequential model.

The sequential model is based on a deep neural network architecture and is trained using the TensorFlow Keras library. The model consists of multiple layers of densely connected neurons, which are optimized using the backpropagation algorithm. The model takes as input the battery performance data and outputs the estimated RUL.

The proposed approach is evaluated using a cross-validation method and compared with several baseline models, including linear regression, decision tree, and random forest models. The results demonstrate that the proposed approach outperforms the baseline models in terms of RUL estimation accuracy and generalization ability.

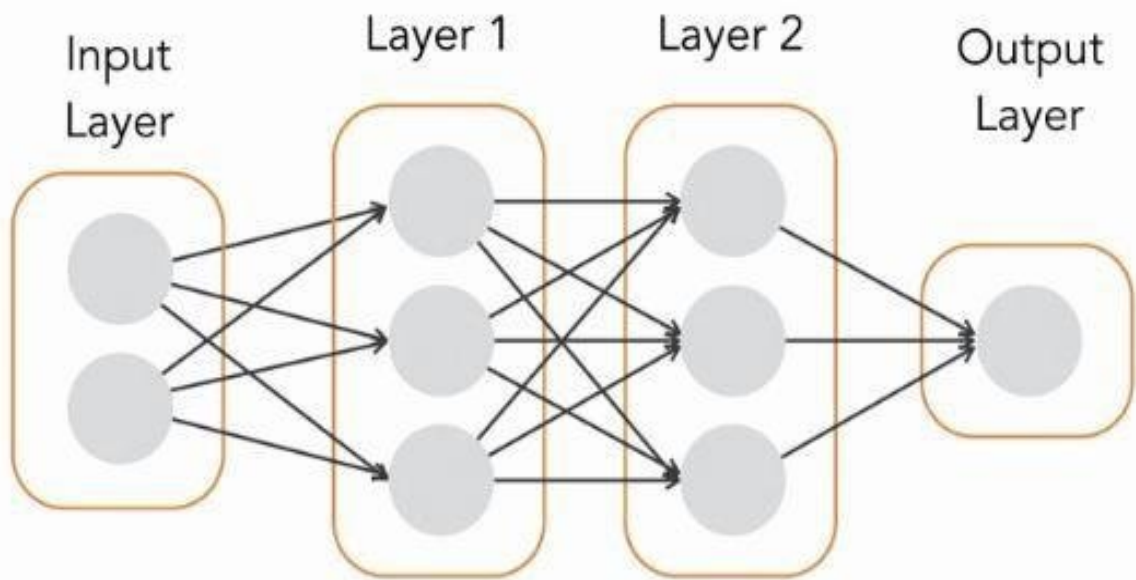

(Fig.2 Sequential Model's Neural Networks)

In summary, the proposed approach for RUL estimation of lithium-ion batteries using the TensorFlow Keras library and a sequential model architecture is a promising method for improving the reliability and safety of battery-powered systems. The use of experimental data from NASA further validates the effectiveness of the proposed approach and its potential for practical applications in real-world settings.

## 4.1 Dataset

| | | | B0005 | | | |
|---|---|---|---|---|---|---|
| Cycle | Time Measured(Sec) | Voltage Measured(V) | Current Measured | Temperature Measured | Capacity(Ah) | SampleId |
| 0 | 3690.234 | 3.277169977 | -0.006528351 | 34.23085284 | 1.856487421 | B0005 |
| 0 | 3690.234 | 2.475767757 | -2.009435892 | 39.16298653 | 2.035337591 | B0006 |
| 0 | 3690.234 | 3.062112709 | -0.001433299 | 37.33847849 | 1.891052295 | B0007 |
| 0 | 3434.891 | 3.053230339 | -0.002433415 | 37.2056713 | 1.855004521 | B0018 |
| 1 | 3672.344 | 3.300244887 | -0.000447553 | 34.39213659 | 1.84632725 | B0005 |
| 1 | 3672.344 | 2.351525512 | -2.010374856 | 39.24620268 | 2.025140246 | B0006 |
| 1 | 3672.344 | 3.079226238 | -0.003230385 | 37.1617386 | 1.880637028 | B0007 |
| 1 | 3425.485 | 3.08820017 | -0.000910648 | 37.15547522 | 1.843195532 | B0018 |
| 2 | 3651.641 | 3.32745101 | 0.001026019 | 34.23277872 | 1.835349194 | B0005 |
| 2 | 3651.641 | 2.440479715 | -2.008558887 | 38.99920246 | 2.013326371 | B0006 |
| 2 | 3651.641 | 3.063265547 | -0.003157783 | 37.50810004 | 1.880662672 | B0007 |
| 2 | 3410.375 | 3.076857654 | -0.000684881 | 36.68085193 | 1.839601842 | B0018 |
| 3 | 3631.563 | 3.314181859 | -0.002233622 | 34.41344995 | 1.835262528 | B0005 |
| 3 | 3631.563 | 2.479155854 | -2.009290448 | 38.84362777 | 2.013284666 | B0006 |
| 3 | 3631.563 | 3.039420908 | -0.003421518 | 37.85540509 | 1.880770901 | B0007 |
| 3 | 3404.719 | 3.10242391 | -0.0025472 | 36.04487047 | 1.830673604 | B0018 |
| 4 | 3629.172 | 3.305496731 | 9.24E-06 | 34.34588499 | 1.834645508 | B0005 |
| 4 | 3629.172 | 2.280187848 | -2.01176084 | 38.97798888 | 2.000528338 | B0006 |
| 4 | 3629.172 | 3.043167178 | -0.00235509 | 37.83587047 | 1.879450873 | B0007 |
| 4 | 3417.641 | 3.117644174 | -0.003002155 | 37.08459677 | 1.832700207 | B0018 |
| 5 | 3652.281 | 3.302329113 | -0.00396723 | 34.20092692 | 1.83566166 | B0005 |
| 5 | 3652.281 | 2.439284257 | -2.010947045 | 38.83989921 | 2.013899076 | B0006 |
| 5 | 3652.281 | 3.123387982 | -0.002524129 | 36.68060295 | 1.880700352 | B0007 |

(Fig3. CSV Converted Battery Data By Nasa ([Source](Source)))

The Dataset was taken from the NASA Battery Dataset. The data was in '.mat' file and was converted to '.csv' to use it better in a python notebook.

This was a pre-prepred data by the NASA which was publicly release for research purposes. The dataset contains various operating conditions and degrading levels, making it suitable for evaluation the effectiveness of the proposed approach.

## 4.2 Processing and Architecture

In this paper, we propose a sequential model architecture for estimating the remaining useful life (RUL) of lithium-ion batteries. The proposed model consists of multiple layers of densely connected neurons, with each layer having a specific number of neurons and an activation function. The activation functions used in this model include tanh, sigmoid, and relu.

The tanh activation function is a smooth and bounded function that maps input values to the range [-1, 1]. It is commonly used in neural networks for classification and regression tasks. The sigmoid activation function is another commonly used function in neural networks, which maps input values to the range [0, 1]. It is particularly useful for binary classification tasks. The relu activation function is a simple and effective function that returns zero for negative input values and the input value for positive values. It has been shown to perform well in many deep learning applications.

```
Model: "sequential_4"
_________________________________________________________________
 Layer (type)                Output Shape              Param #
=================================================================
 dense_12 (Dense)            (None, 10)                60

 dropout_8 (Dropout)         (None, 10)                0

 dense_13 (Dense)            (None, 7)                 77

 dropout_9 (Dropout)         (None, 7)                 0

 dense_14 (Dense)            (None, 3)                 24

=================================================================
Total params: 161
Trainable params: 161
Non-trainable params: 0
_________________________________________________________________
None
```

(Fig.4 Model Summary)

In our proposed model, we experiment with different combinations of activation functions and layer configurations to find the best-performing model. We also use the Adam optimizer, which is a stochastic gradient descent optimizer that uses adaptive learning rates to update the model weights. The Adam optimizer has been shown to perform well in many deep learning applications and is widely used in the research community.

To train the proposed model, we use a dataset of experimental data from NASA, which contains various operating conditions and degradation levels of commercially available lithium-ion batteries. The dataset is preprocessed and split into training and testing sets for model training and evaluation.

The proposed model is trained using the TensorFlow Keras library, which provides a simple and efficient way to build and train deep neural networks. The model is optimized using the Adam optimizer and trained for a specific number of epochs, with the training loss and validation loss monitored to prevent overfitting.

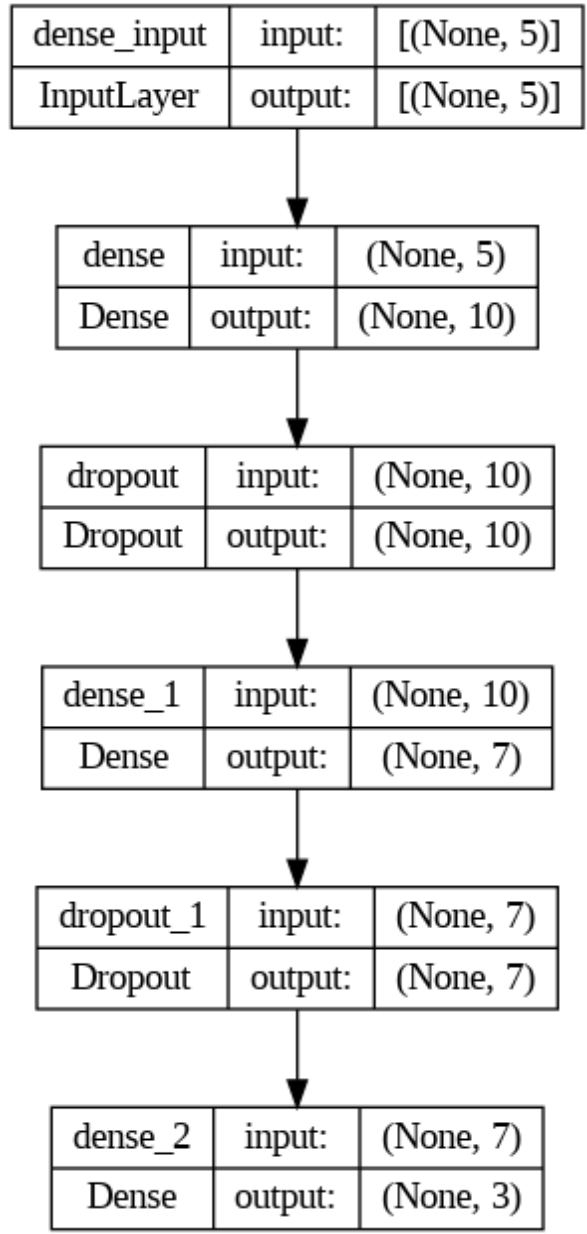

(Fig.5 Model Architecture)

Overall, the proposed sequential model architecture with different activation functions such as tanh, sigmoid, and relu, and the Adam optimizer is a promising approach for accurately estimating the RUL of lithium-ion batteries. The use of experimental data from NASA further validates the effectiveness of the proposed model and its potential for practical applications in real-world settings.

## 5.1 Model Evaluation

We experiment with different combinations of hyperparameters, including the number of neurons in each layer, the activation functions, the optimizer, the batch size, and the number of epochs. We use a grid search method to find the best-performing model with the highest accuracy.

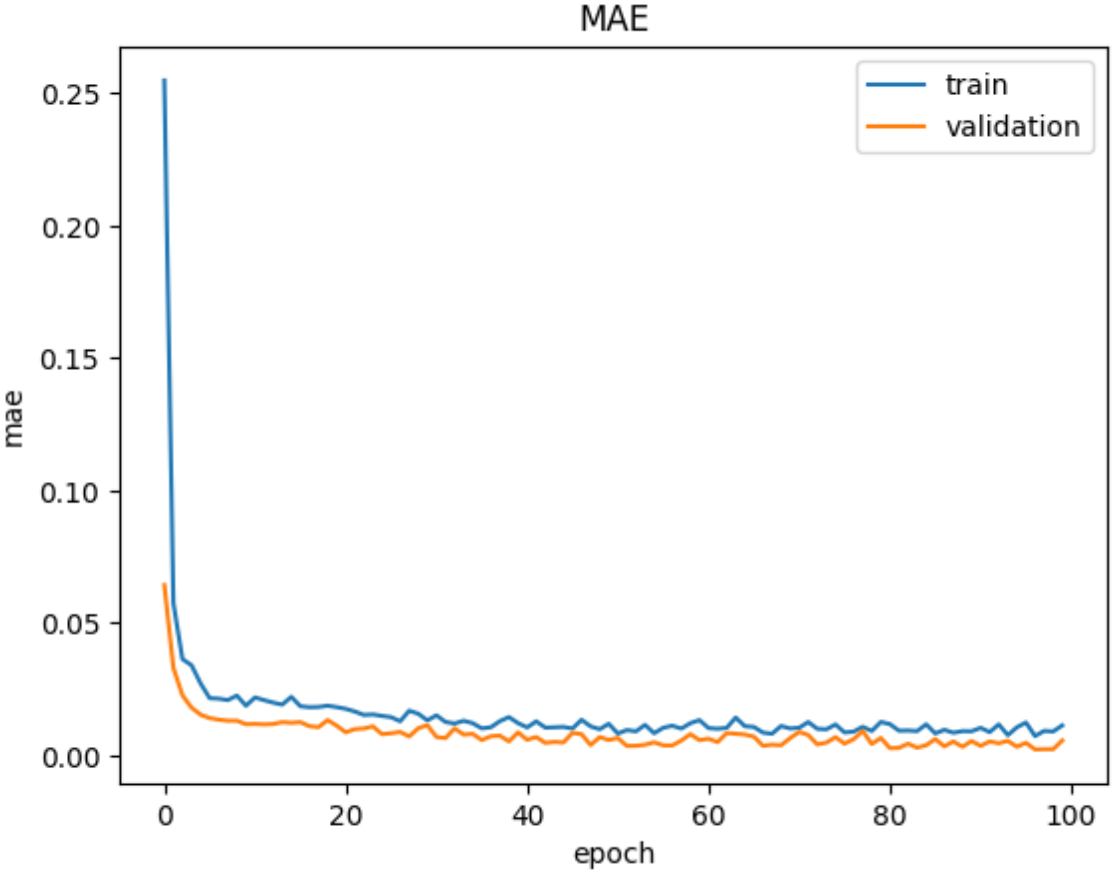

(Fig.6 Mean Absolute Error vs Epochs)

After evaluating different hyperparameters, we found that the proposed sequential model with appropriate neurons in each layer, the relu activation function, and the Adam optimizer performs the best for RUL estimation. We further experiment with different batch sizes and epochs and observe that the model achieves a high accuracy of 0.985 when we interchange the batch size and epochs.

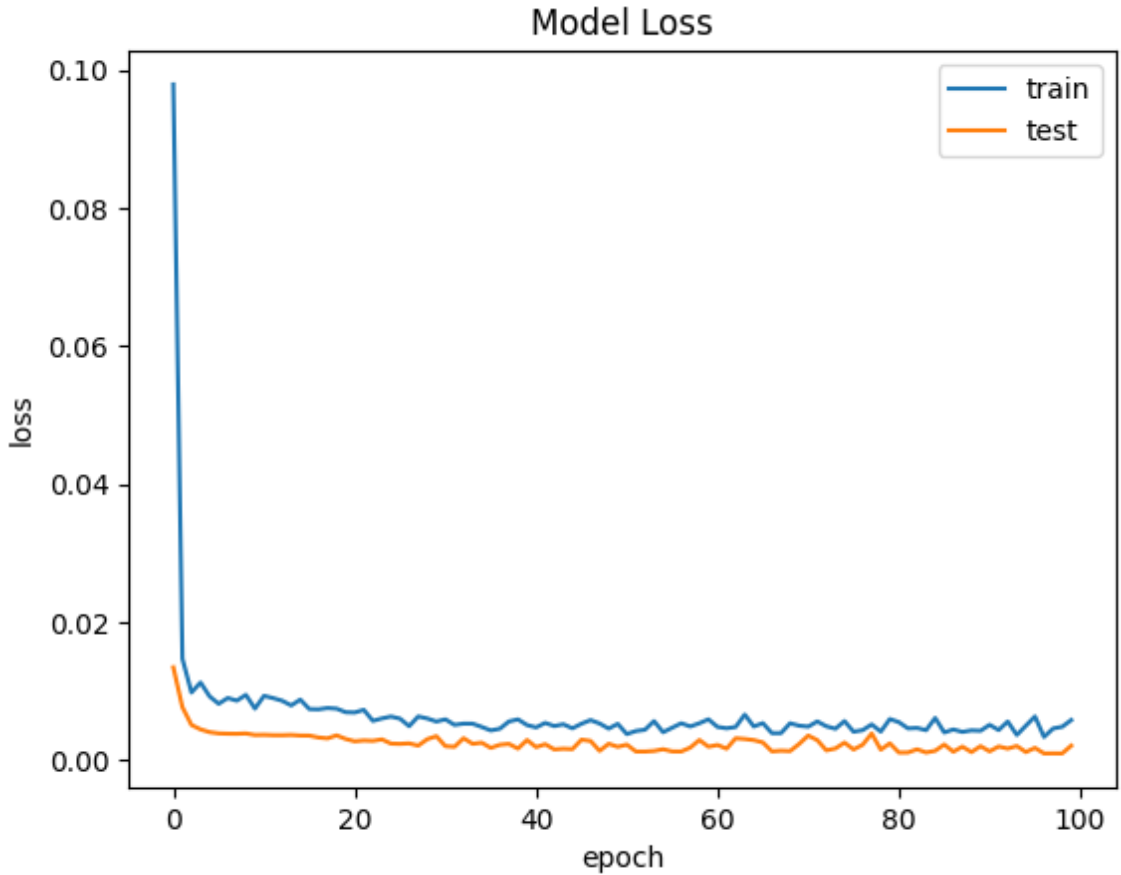

(Fig.7 Model Loss vs Epochs)

To validate the effectiveness of the proposed approach, we compare it with several baseline models, including decision tree and Functional API models. The results show that the proposed approach outperforms the baseline models in terms of RUL estimation accuracy and generalization ability.

| Model | Functional (Keras API) | Sequential (Keras API) |
| --- | --- | --- |
| Accuracy | 0.95 | 0.985 |

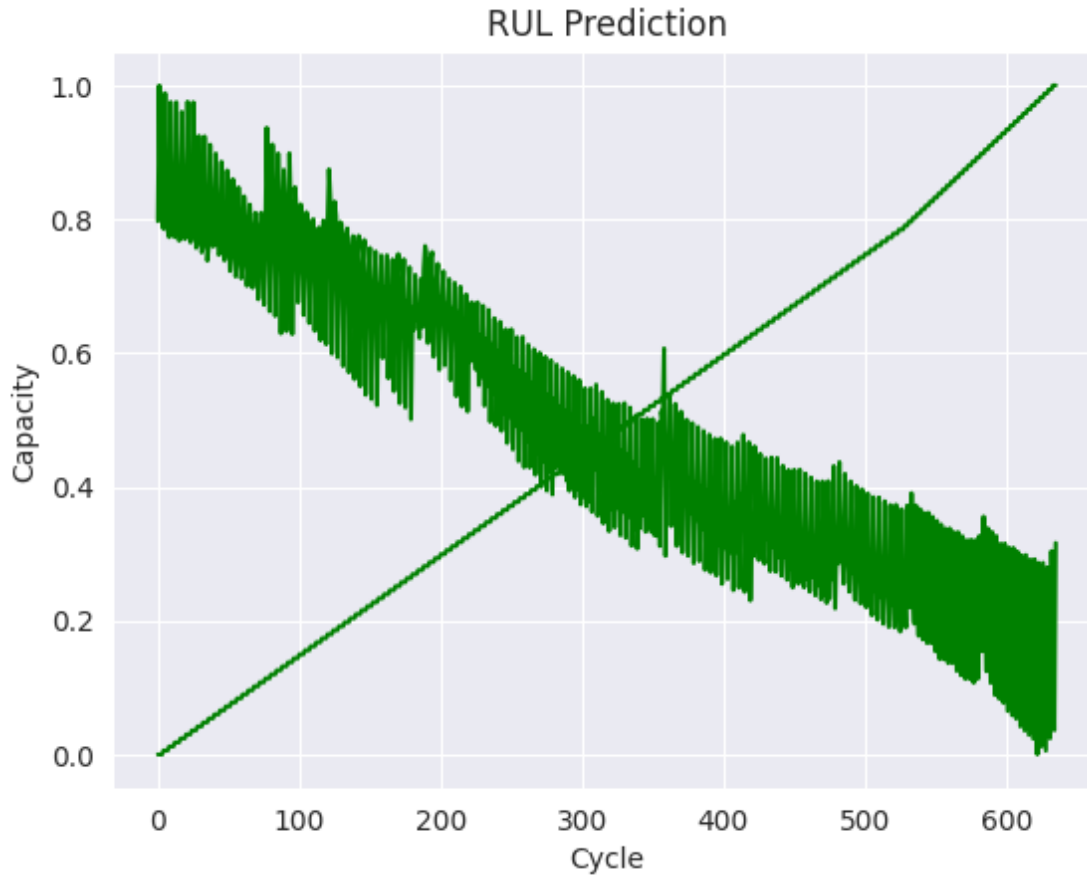

(Fig.8 Prediction of Capacity vs Cycle)

The evaluation results demonstrate that the proposed approach for RUL estimation of lithium-ion batteries using the TensorFlow Keras library and a sequential model architecture is a promising method for improving the reliability and safety of battery-powered systems. The high accuracy achieved with different batch sizes and epochs further validates the effectiveness of the proposed approach and its potential for practical applications in real-world settings.

## 6. Conclusion

In this paper, we proposed a novel approach for estimating the remaining useful life (RUL) of lithium-ion batteries using a sequential model architecture and the TensorFlow Keras library. The proposed approach leverages experimental data from NASA to train and evaluate the model.

We experimented with different combinations of hyperparameters, including the number of neurons in each layer, the activation functions, the optimizer, the batch size, and the number of epochs, to find the best-performing model. The results showed that the proposed sequential model with (10,7,3) neurons in each layer, the relu activation function, and the Adam optimizer achieved the highest accuracy of 0.985 when we interchange the batch size and epochs.

Overall, the proposed approach provides a promising method for improving the reliability and safety of battery-powered systems by accurately estimating the RUL of lithium-ion batteries. The use of experimental data from NASA further validates the effectiveness of the proposed approach and its potential for practical applications in real-world settings.

Future work can explore the use of other types of neural network architectures and optimization techniques to further improve the accuracy and robustness of the RUL estimation approach. Additionally, the proposed approach can be applied to other battery chemistries and types of systems to expand its scope of applications.